\documentclass[10pt,twocolumn,letterpaper]{article}

\usepackage{wacv}              % To produce the CAMERA-READY version
\usepackage{graphicx}
\usepackage{booktabs}
\usepackage{wrapfig,lipsum,booktabs}
\usepackage{multirow}
\usepackage{threeparttable}
\usepackage{bm}
\usepackage{algorithmicx}
\usepackage{algorithm}% http://ctan.org/pkg/algorithms
\usepackage{algpseudocode}% http://ctan.org/pkg/algorithmicx
\usepackage{amsmath}

\makeatletter
\def\blfootnote{\xdef\@thefnmark{}\@footnotetext}
\makeatother

\usepackage[pagebackref,breaklinks,colorlinks]{hyperref}

\usepackage[capitalize]{cleveref}
\crefname{section}{Sec.}{Secs.}
\Crefname{section}{Section}{Sections}
\Crefname{table}{Table}{Tables}
\crefname{table}{Tab.}{Tabs.}

\def\wacvPaperID{315} % *** Enter the WACV Paper ID here
\def\confName{WACV}
\def\confYear{2025}

\DeclareMathOperator*{\argmin}{arg\,min}
\newcommand{\OurMethod}{\emph{FastVideoEdit}}

\begin{document}

%%%%%%%%% TITLE - PLEASE UPDATE
\title{FastVideoEdit: Leveraging Consistency Models for Efficient Text-to-Video Editing}

% \author{First Author\\
% Institution1\\
% Institution1 address\\
% {\tt\small firstauthor@i1.org}
% % For a paper whose authors are all at the same institution,
% % omit the following lines up until the closing ``}''.
% % Additional authors and addresses can be added with ``\and'',
% % just like the second author.
% % To save space, use either the email address or home page, not both
% \and
% Second Author\\
% Institution2\\
% First line of institution2 address\\
% {\tt\small secondauthor@i2.org}
% }

\author{
    Youyuan Zhang\textsuperscript{\rm 1} \quad
    Xuan Ju\textsuperscript{\rm 2}\thanks{Corresponding authors.} \quad
    James J. Clark\textsuperscript{\rm 1}\footnotemark[1]
    \\
    \textsuperscript{\rm 1}McGill University \quad
    \textsuperscript{\rm 2}The Chinese University of Hong Kong
    \\
    {\tt\small youyuan.zhang@mail.mcgill.ca, xju22@cse.cuhk.edu.hk, james.clark1@mcgill.ca}
}

\maketitle

%%%%%%%%% ABSTRACT
\begin{abstract}
   Diffusion models have demonstrated remarkable capabilities in text-to-image and text-to-video generation, opening up possibilities for video editing based on textual input. However, the computational cost associated with sequential sampling in diffusion models poses challenges for efficient video editing. Existing approaches relying on image generation models for video editing suffer from time-consuming one-shot fine-tuning, additional condition extraction, or DDIM inversion, making real-time applications impractical. In this work, we propose \OurMethod, an efficient zero-shot video editing approach inspired by Consistency Models (CMs). By leveraging the self-consistency property of CMs, we eliminate the need for time-consuming inversion or additional condition extraction, reducing editing time. Our method enables direct mapping from source video to target video with strong preservation ability utilizing a special variance schedule. This results in improved speed advantages, as fewer sampling steps can be used while maintaining comparable generation quality. Experimental results validate the state-of-the-art performance and speed advantages of \OurMethod~across evaluation metrics encompassing editing speed, temporal consistency, and text-video alignment.
\end{abstract}

\section{Introduction}
\label{sec:introduction}

\begin{figure}[htbp]
    \centering
    \includegraphics[width=0.99\linewidth]{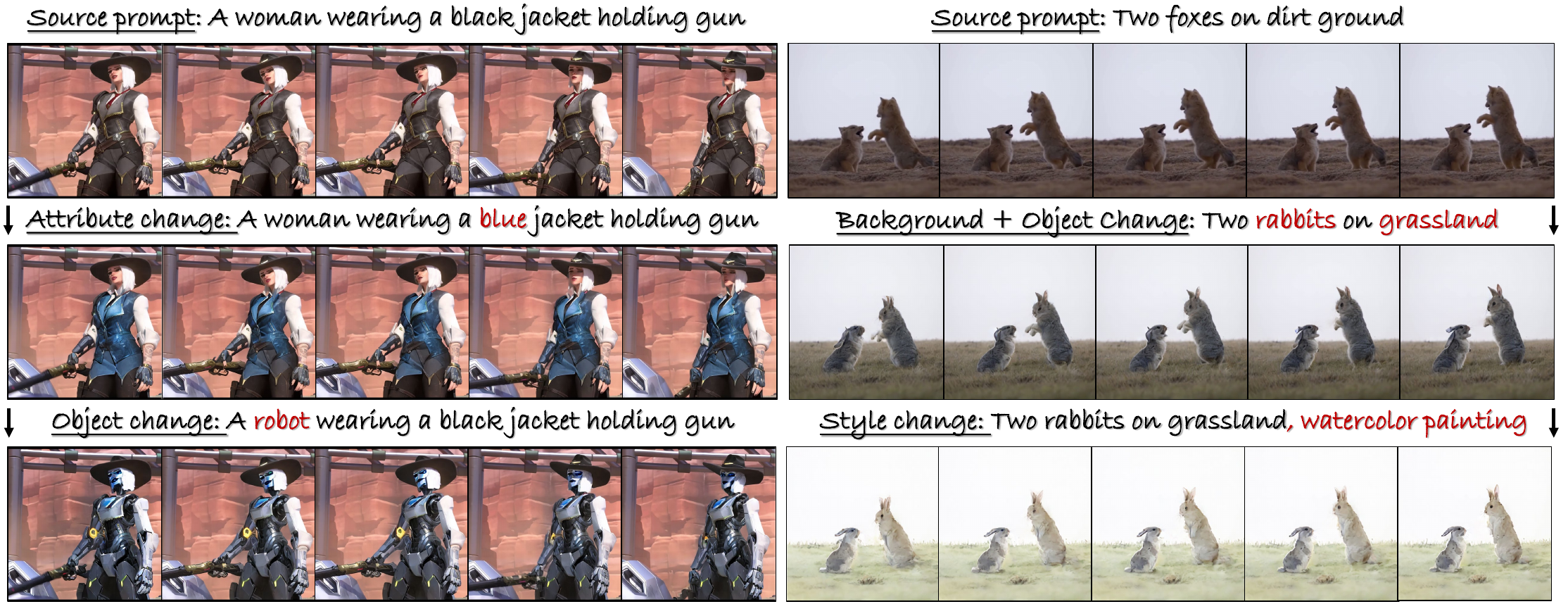}
    \caption{\textbf{Editing Results of \OurMethod.} \OurMethod~offers efficient, consistent, high-quality, and text-aligned editing capabilities for both artificial (left col) and natural (right col) videos. The top row displays the source video, while the second and third rows showcase two edited videos. Each row features a text prompt at the top, with the edited words highlighted in red. This visual representation effectively demonstrates how our method can successfully achieve desired edits such as attribute change, object change, background change, and style change.
    }
    \label{fig:teaser}
\end{figure}

Diffusion models~\cite{ho2020denoising,song2020denoising,ho2022imagen,balaji2022ediffi} have gained significant attention due to their remarkable capabilities in text-to-image~\cite{rombach2022high,song2020denoising,ho2020denoising} and text-to-video generation~\cite{ho2022imagen,singer2022make,blattmann2023align,gupta2023photorealistic,videoworldsimulators2024}.
Leveraging the capabilities of these models, it becomes feasible to manipulate videos~\cite{videoworldsimulators2024} based on textual input, holding great potential for various applications in areas such as film production and content creation.

However, the computational cost associated with sequential sampling in diffusion models presents a significant challenge for efficient inference, especially in video editing scenarios where a set of frames need to be processed. 
Moreover, the absence of high-quality open-source video diffusion models~\cite{esser2023structure,molad2023dreamix} that can generate consistent editing results within a single test time inference, combined with the constraints on video duration of video diffusion models, has led to the adoption of existing image generation models for achieving accurate video editing~\cite{bar2022text2live,qi2023fatezero,geyer2023tokenflow}.
To align the distribution between image and video models and perform accurate video editing, some methods employ a test-time one-shot fine-tuning for inflated image generation model on each input video~\cite{wu2023tune,shin2023edit,liu2023video,wang2023zero,ma2023magicstick}. 
However, this process further exacerbates the time-consuming nature of the editing process, which makes it impractical for real-time applications. 

To enable faster video editing, three types of zero-shot methods have been proposed in the literature: (1) \textit{Layer-atlas-based methods}~\cite{bar2022text2live,kasten2021layered,chai2023stablevideo}, which involve editing the video on a flattened texture map and ensuring the temporal consistency by guaranteeing texture map consistency. However, the absence of a 3D motion prior in the 2D atlas approach results in suboptimal performance. (2) \textit{Dual-branch methods}~\cite{qi2023fatezero,ceylan2023pix2video,geyer2023tokenflow,cong2023flatten}, which leverage \textit{Denoising Diffusion Implicit Models} (DDIM)~\cite{song2020denoising} to extract source video features and generate novel content based on the target diffusion branch. The use of DDIM inversion leads to a doubling of the inference time required for video editing. (3) \textit{Additional conditional constraints incorporating methods}~\cite{yang2023rerender,wang2023zero,chen2023control,zhang2023controlvideo}, which involve directly adding noise to the source video and denoising the noisy video using a conditioned diffusion model for preserving essential content while imposing restrictions on the editing process. While these methods are efficient during diffusion model inference, they do require additional information extraction, which slows down the overall speed of the process.

To address the issue of long computational times encountered in previous video editing methods, we introduce \OurMethod, which is inspired by recent advances in \textit{Consistency Models} (CMs)~\cite{song2023consistency}. 
Specifically, \OurMethod~is a zero-shot video editing approach that not only achieves state-of-the-art performance but also significantly reduces editing time by eliminating the need for time-consuming inversion or additional condition extraction steps. 
The key insight of our proposed method is that the self-consistency property of CMs enables a special variance schedule that facilitates the editing process, transforming it from a process of adding noise and then denoising to one of a direct mapping from source video to target video.
Furthermore, the content preservation capability of CMs enables the use of fewer sampling steps while maintaining comparable generation quality, which results in an improved speed advantage of \OurMethod.

To evaluate \OurMethod, we consider metrics that encompass editing speed, temporal consistency, and text-video alignment. We compare the performance of \OurMethod~with previous video editing methods using the TGVE 2023 open-source dataset~\cite{wu2023cvpr} as our benchmark.
The results demonstrate the superior performance of \OurMethod~in terms of editing quality. 
Additionally, \OurMethod~achieves this superior performance while requiring significantly less time for video editing tasks. 
This shows the efficiency and effectiveness of our approach, making it a standout choice for efficient high-quality video editing.

\section{Related Work}
\label{sec:related_work}

\subsection{Video Editing with Diffusion Models}

The remarkable success of diffusion-based text-to-image~\cite{rombach2022high,song2020denoising,ho2020denoising} and text-to-video generation models~\cite{ho2022imagen,singer2022make,blattmann2023align,gupta2023photorealistic,videoworldsimulators2024} has opened up new possibilities for exciting opportunities in text-based image~\cite{hertz2022prompt,ju2023direct} and video editing~\cite{esser2023structure}.
Although editing video directly through video diffusion models~\cite{esser2023structure,molad2023dreamix,esser2023structure} show high temporal consistency, the challenges associated with extensive video model training, unstable generation quality, and video duration time limit make using inflated off-the-shelf image generation models a preferable choice for video editing, which inflating 2D model to 3D with an additional temporal channel. 

Specifically, several works require a test-time one-shot fine-tuning on the inflated image generation model with each input video~\cite{wu2023tune,shin2023edit,liu2023video,wang2023zero,ma2023magicstick}, which is time-consuming and too slow for real-time applications. 
Zero-shot video editing methods~\cite{bar2022text2live,kasten2021layered,chai2023stablevideo,qi2023fatezero,hertz2022prompt,wang2023zero,ceylan2023pix2video,yang2023rerender,geyer2023tokenflow,zhang2023controlvideo,cong2023flatten,chen2023control} leverage training-free editing techniques with specialized modules to enhance temporal consistency across frames, which provide a practical and efficient solution for editing videos without the need of extensive training.
Specifically, layer-atlas-based methods~\cite{bar2022text2live,kasten2021layered,chai2023stablevideo} edit the video on a flattened texture map, however the lack of 3d motion prior in 2d atlas leads to suboptimal performance.
FateZero~\cite{qi2023fatezero} solves this problem with a two-branch inflated image diffusion model that merges attention features of the structural preservation branch and editing branch. 
Similarly, Text2Video-Zero~\cite{khachatryan2023text2video} and Pix2Video~\cite{ceylan2023pix2video} align the feature of the source image and target image via an attention operation.
To enhance pixel-level temporal consistency, Rerender A Video~\cite{yang2023rerender}, TokenFlow~\cite{geyer2023tokenflow}, and Flatten~\cite{cong2023flatten} extract temporal-aware inter-frame features to propagate the edits throughout the video.
% ControlVideo~\cite{zhang2023controlvideo} and Control-A-Video~\cite{chen2023control} use ControlNet~\cite{zhang2023adding} to help editing.
However, previous zero-shot methods that relied on flattened image diffusion were limited by the need for DDIM inversion or additional conditional constraints (\textit{e.g.}, optical flow), resulting in a long runtime. 
In contrast, our proposed \OurMethod~directly incorporates editing into the inference process by leveraging consistency models~\cite{song2023consistency}, which ensures both runtime efficiency and effective modifications.

\subsection{Efficient Diffusion Models}

To tackle the computational time limitations of diffusion models caused by the sequential sampling strategy, faster numerical ODE solvers~\cite{song2020denoising,zhang2022fast,lu2022dpm} or distillation techniques~\cite{luhman2021knowledge,salimans2022progressive,meng2023distillation,zheng2023fast} have been employed. 
While these methods can be integrated into existing diffusion-based video editing techniques, they still face the challenge of requiring DDIM inversion or additional conditional constraints for essential content preservation.

Recently, the introduction of \textit{Consistency Models} (CMs)~\cite{song2023consistency,xu2023infedit} has enabled faster generation by sampling along a trajectory map, thereby opening up exciting possibilities for more efficient video editing techniques. The few-step sampling strategy is particularly suitable for efficient video editing with a fast sampling speed and strong reconstruction ability. \OurMethod~leverages the self-consistency characteristic of CMs, where the improved essential content preservation ability eliminates the need for accurate DDIM inversion and additional conditional constraints. Concurrent to our approach, OCD~\cite{kahatapitiya2024object} separates diffusion sampling for edited objects and background areas, focusing most denoising steps on the former to enhance efficiency. \OurMethod~can be directly combined with OCD to further enhance the overall efficiency of video editing.

\section{Preliminaries}
\label{sec:preliminaries_and_motivation}

Diffusion models include a forward process that adds Gaussian noise $\epsilon$ to convert clean sample $z_0$ to noise sample $z_T$, 
and a backward process that iteratively performs denoising from $z_T$ to $z_0$, where $T$ represents the total number of timesteps.
The denoising process of DDPM sampling~\cite{ho2020denoising} at step $t$ can be formulated as:

\vspace{-0.4cm}
\begin{equation}
\label{eq:fwd}
\begin{aligned}
    z_{t-1}
    &= \sqrt{{{\alpha}}_{t-1}} \left( \frac{z_t - \sqrt{1 - {{\alpha}}_t} \varepsilon_{\theta}(z_t, t)}{\sqrt{{\alpha}}_t} \right) && \text{(predicted $z_0$)} \\
    & + \sqrt{1 - {\alpha}_{t-1} - \sigma_t^2} \cdot \varepsilon_{\theta}(z_t,t) && \text{(direction to $z_t$)} \\
    & + \sigma_t \varepsilon_t\quad \text{where } \varepsilon_t \sim \mathcal{N}(\bm{0},\bm{I}) && \text{(random noise).}
\end{aligned}
\end{equation}

By setting $\sigma_t$ to zero, DDIM sampling~\cite{song2020denoising} results in an implicit probabilistic model with a deterministic forward process:

\vspace{-0.3cm}
\begin{equation}
\label{eq:cond}
    \bar{z}_0 
    = f_\theta(z_t, t) 
    = \left(z_t - \sqrt{1 - {\alpha}_t} \cdot \varepsilon_\theta(z_t, t) \right) / \sqrt{{\alpha}_t}.
\end{equation}

Following DDIM, we can use the function $f_\theta$ to predict and reconstruct $\bar{z_0}$ given noise sample $z_t$, where $t\sim \left[ 1,T \right]$, $\alpha$ is the hyper-parameter, $\varepsilon_{\theta}$ is a learnable network, and $T$ represents the total number of timesteps.

Sampling in CMs~\cite{song2023consistency} is carried out through a sequence of timesteps $\tau_{1:n} \in [t_0,T]$. 
Starting from an initial noise $\hat{z}_T$ and $z_0^{(T)} = f_\theta(\hat{z}_T, T)$, at each time-step $\tau_{i}$, the process samples $\varepsilon \sim \mathcal{N}(\bm{0}, \bm{I})$ and iteratively updates the \textit{Multistep Consistency Sampling} process through the following equation:

\vspace{-0.3cm}
\begin{equation}
\begin{aligned}
\label{eq:mscs}
    \hat{z}_{\tau_i} &= z_0^{(\tau_{i+1})} + \sqrt{\tau_i^2 - t_0^2 }\varepsilon \\
    z_0^{(\tau_{i})} &= f_{\theta}(\hat{z}_{\tau_i}, \tau_i).
\end{aligned}
\end{equation}

When combined with a condition $c$ with classifier-free guidance~\cite{ho2022classifier}, sampling in CMs at $\tau_{i}$ starts with $\varepsilon \sim \mathcal{N}(\bm{0}, \bm{I})$ and updates through:

\vspace{-0.3cm}
\begin{equation}
\begin{aligned}
\label{eq:lcm-samp}
    \hat{z}_{\tau_i} &= \sqrt{{\alpha}_{\tau_i}}z_0^{(\tau_{i+1})} + \sigma_{\tau_i} \varepsilon, \\
    z_0^{(\tau_{i})} &= f_{\theta}(\hat{z}_{\tau_i}, \tau_i, c).
\end{aligned}
\end{equation}

\label{prop:const}
Consider a special case of Eq.~\ref{eq:fwd} where $\sigma_t$ is chosen as $\sqrt{1 - \alpha_{t-1}}$ at all times $t$. Then the DDPM forward process naturally aligns with the Multistep Consistency Sampling, and the second term of Eq.~\ref{eq:fwd} vanishes:

\vspace{-0.3cm}
\begin{equation}
\label{eq:ddcm1}
\begin{aligned}
    z_{t-1}
    &= \sqrt{{{\alpha}}_{t-1}} \left( \frac{z_t - \sqrt{1 - {{\alpha}}_t} \varepsilon_{\theta}(z_t, t)}{\sqrt{{\alpha}}_t} \right) && \text{(predicted $z_0$)} \\
    & + \sqrt{1 - \alpha_{t-1}} \varepsilon_t\quad \varepsilon_t \sim \mathcal{N}(\bm{0},\bm{I}) && \text{(random noise).}
\end{aligned}
\end{equation}

\noindent
Consider $f(z_t, t; z_0) = \left( z_t - \sqrt{1 - {\alpha}_t} \varepsilon'(z_t,t;z_0) \right) / \sqrt{{\alpha}_t}$, where the initial $z_0$ is available
% (which is the case for image editing applications) 
and we replace the parameterized noise predictor  $\varepsilon_\theta$ with $\varepsilon'$ more generally.
Eq.~\ref{eq:ddcm1} turns into the following expression:

\vspace{-0.3cm}
\begin{equation}
\begin{aligned}
    z_{t-1} = \sqrt{{{\alpha}}_{t-1}} f(z_t,t;z_0) + \sqrt{1 - \alpha_{t-1}} \varepsilon_t
\end{aligned}
\end{equation}

\noindent
which is in the same form as the Multistep Consistency Sampling step in Eq~\ref{eq:lcm-samp}.

In order to make $f(z_t, t)$ self-consistent so that it can be considered as a consistency function, i.e., $f(z_t, t; z_0) = z_0$, we can directly solve the equation and $\varepsilon'$ can be computed without parameterization:

\vspace{-0.3cm}
\begin{equation}
\label{eq:ddcm-eps}
    \varepsilon^{\text{cons}} = \varepsilon'(z_t, t;z_0) = \frac{z_t - \sqrt{{\alpha}_t} z_0}{\sqrt{1 - {\alpha}_t}}.
\end{equation}

We arrive at a non-Markovian forward process, in which $z_t$ directly points to the ground truth $z_0$ without neural prediction, and $z_{t-1}$ does not depend on the previous step $z_t$ like a consistency model. 

\section{Method}
\label{sec:method}

\begin{figure*}[htbp]
    \centering
    \begin{subfigure}{.89\linewidth}
        \centering
        \includegraphics[width=0.9\linewidth]{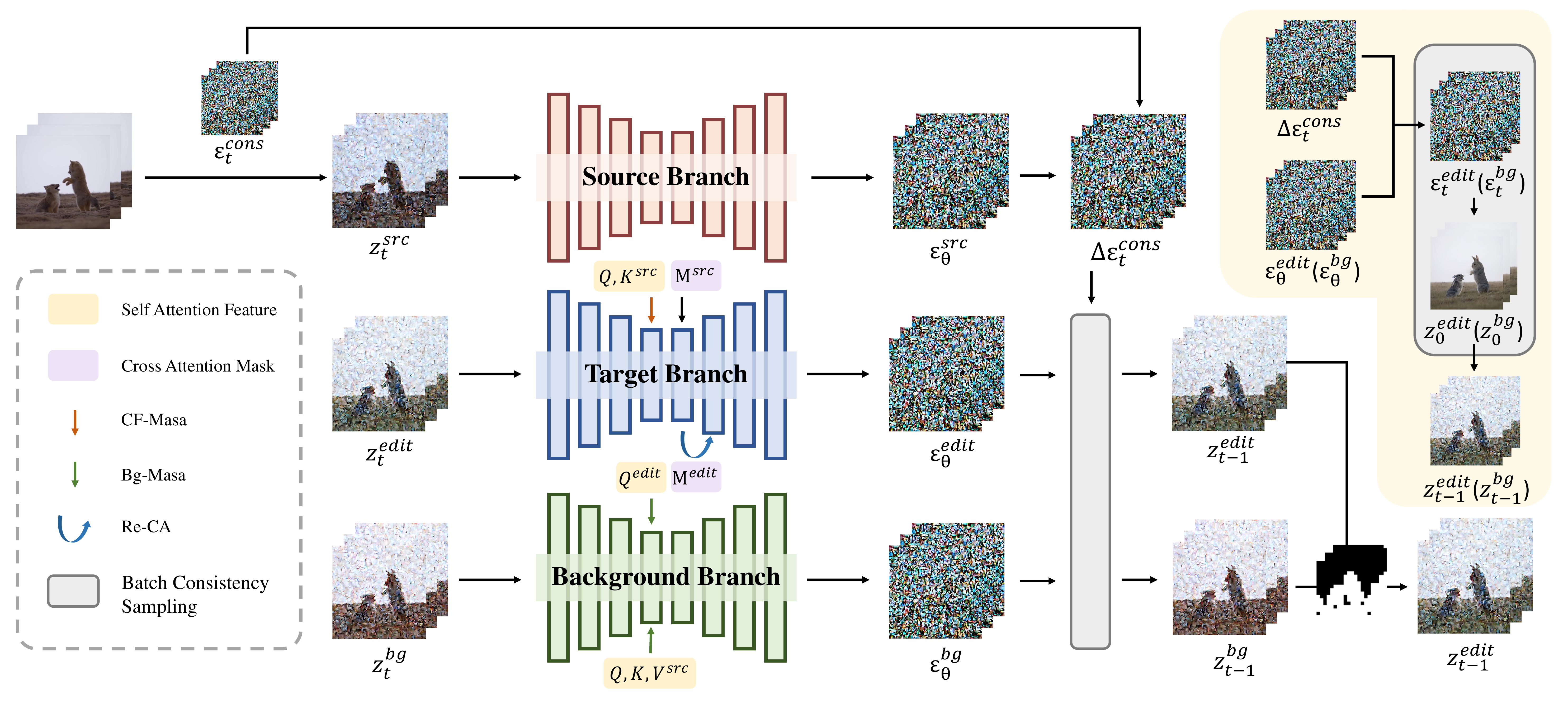}
    \end{subfigure}
    \caption{\textbf{Overview of \OurMethod.} Our model directly denoises three branches of batch frames using three attention control methods: $\text{CF-Masa}$, $\text{Re-CA}$ and $\text{Bg-Masa}$. The model uses batch consistency sampling (\text{BCS}) with LCMs to improve efficiency, background latent blending to align editing content with source video and TokenFlow propagation to further improve temporal consistency. The right shaded part elaborates detailed operation of using batch consistency sampling to estimate noise in editing branch and background branch.
    }
    \label{fig:overview}
\end{figure*}

The task of video editing can be described as the following: Given an ordered set of $m$ source video frames $\mathcal{I}_\text{src} = \{I_1^\text{src}, I_2^\text{src}, ..., I_m^\text{src}\}$ and a source prompt $\mathcal{P}_\text{src}$ describing the source video, we aim to generate an edited video with temporally consistent frames $\mathcal{I}_\text{edit} = \{I_1^\text{edit}, I_2^\text{edit}, ..., I_m^\text{edit}\}$ according to a target prompt $\mathcal{P}_\text{tgt}$.

This paper introduces \OurMethod, an end-to-end video edit framework that edits video efficiently while producing high-quality and temporally consistent editing content. 
Notably, our method achieves better background preservation compared with existing methods when editing foreground object-level attributes. Unlike many existing methods that depend on additional estimations such as depth control, edge control, or optical flow, \OurMethod~requires only the source video frames and prompts as input throughout the editing process.

\subsection{Video Reconstruction with Consistency Model}

To our knowledge, \OurMethod~is the first video editing framework that eliminates the need of the DDIM inversion process while simultaneously performing a complete denoising process on individual video frames. To enable direct editing of the source video without the need for the inversion process, we leverage a consistency model inspired by InfEdit~\cite{xu2023infedit}. The key idea to reconstruct source latent is to start with randomly sampled reconstruction noise rather than randomly initialized noisy latents. Following the \textit{Multistep Consistency Sampling} in Eq~\ref{eq:mscs}, we sample a noise $\varepsilon_t^{\text{cons}}$ at each timestep $t$ and the noisy latent $z_t^{\text{src}}$ becomes directly tractable when $z_0^{\text{src}}$ is given in the editing problem. Instead of denoising the randomly initialized noisy latent $z_T^{\text{src}}$, the whole trajectory of $\{z_t^{\text{src}}\}$ is obtained directly from the sampled noise trajectory $\{\varepsilon_t^{\text{cons}}\}$, and in the reverse direction each $\varepsilon_t^{\text{cons}}$ can be used to reconstruct $z_0^{\text{src}}$ given $z_t^{\text{src}}$. The mappings between $z_t^{\text{src}}$ and $\{\varepsilon_t^{\text{cons}}\}$ given $z_0^{\text{src}}$ are given by:

\vspace{-0.3cm}
\begin{equation}
\label{eq:src}
\begin{aligned}
    z_t^{\text{src}} &= \sqrt{{\alpha}_t} z_0^{\text{src}} + \sqrt{1-{\alpha}_t} \varepsilon_t^{\text{cons}} \\
    \varepsilon^{\text{cons}}_t &= (z_t^{\text{src}} - \sqrt{{\alpha}_t} z_0^{\text{src}})/ \sqrt{1-{\alpha}_t},
\end{aligned}
\end{equation}

\noindent
where $\varepsilon_t^{\text{cons}} \sim \mathcal{N}(\bm{0},\bm{I})$ is sampled independently at each timestep. As a result, the reconstructed latent $z_t = z_0$ is guaranteed at each timestep using Eq (\ref{eq:cond}).

\subsection{Video Editing with Consistency Model}
\label{sec:video_editing_with_consistency_model}

This section introduces the method to compute $z_0^\text{edit}$ given $z_0^{\text{src}}$. In addition to $z_t^{\text{src}}$ and $\varepsilon^{\text{cons}}_t$ obtained from Eq (\ref{eq:src}), we need to predict the editing noise $\varepsilon_\theta(z_t^{\text{edit}}, t, \mathcal{P}_{\text{tgt}})$ to generate the editing latent $z_0^\text{edit}$ according to target prompt $\mathcal{P}_{\text{tgt}}$. Due to the self-consistency property of LCMs, the gap between $\varepsilon_\theta(z_t^{\text{edit}}, t, \mathcal{P}_{\text{tgt}})$ and $\varepsilon_t^{\text{edit}}$ is small. Therefore, using the noise calibration $\Delta\varepsilon_t^{\text{cons}}$ from $\varepsilon_\theta(z_t^{\text{src}}, t, \mathcal{P}_{\text{src}})$ to the ground-truth source reconstruction noise $\varepsilon^{\text{cons}}_t$, we can estimate the editing reconstruction noise as well as the editing latent $z_0^{\text{edit}}$ at each timestep $t$:

\vspace{-0.2cm}
\begin{equation}
\label{eq:edit}
\begin{aligned}
    \Delta\varepsilon_t^{\text{cons}} &= \varepsilon^{\text{cons}}_t - \varepsilon_\theta(z_t^{\text{src}}, t, \mathcal{P}_s) \\
    \varepsilon_t^{\text{edit}} &= \varepsilon_\theta(z_t^{\text{edit}}, t, \mathcal{P}_t) + \Delta\varepsilon_t^{\text{cons}} \\
    z_0^{\text{edit}} &= \left(z_t^{\text{edit}} - \sqrt{1 - {\alpha}_t} \cdot \varepsilon_t^{\text{edit}} \right) / \sqrt{{\alpha}_t}.
\end{aligned}
\end{equation}

Compared with editing a single frame, we impose the constraints that the initial latent and random noise sampled at each timestep are identical across all frames. Since the forward process of the denoising network $\varepsilon_\theta(\cdot, \cdot, \cdot)$ as well as the calibration process of noise and the updating process of latent are all deterministic relative to their inputs, identical initial latents and noise samples at each timestep result in identical output latents when source latents are also identical. In practice, if source latents are temporally consistent and close to each other, the output latents should also maintain good temporal consistency.

\subsection{Batch Attention Control}

As an end-to-end inference-based editing framework \OurMethod~starts with directly denoising the batched latent $\mathcal{Z}_t^{\text{edit}}$ according to the target prompt $\mathcal{P}_{\text{tgt}}$. A naive way of editing the target frame latent $z_0^{\text{src}}$ by the target prompt is to denoise the DDIM inversion $z_T^{\text{inv}}$ of $z_0^{\text{src}}$ iteratively through $\varepsilon_\theta(z_t^{\text{inv}}, t, \mathcal{P}_{\text{tgt}})$. In section \ref{sec:video_editing_with_consistency_model}, we introduced consistency model-based batch editing which leverages the property of LCMs to skip the time-consuming DDIM inversion process and directly denoise randomly initialized latent while keeping content aligned faithfully with source frames. However, without additional control, denoising conditioned on a target prompt $\mathcal{P}_{\text{tgt}}$ can still produce editing content distinct from the source content.

Inspired by MasaCtrl~\cite{cao2023masactrl} and Prompt-to-prompt~\cite{hertz2022prompt}, we propose \textit{Cross-Frame Mutual Self-Attention} (CF-Masa) and \textit{Re-weighted Cross Attention} (Re-CA) to allow further attention control when denoising the $z_t^{\text{edit}}$ conditioned on $\mathcal{P}_{\text{tgt}}$. Specifically, we concurrently denoise two batched latents $[\mathcal{Z}_t^{\text{src}}, \mathcal{Z}_t^{\text{edit}}]$ conditioned on $[\mathcal{P}_{\text{src}}, \mathcal{P}_{\text{tgt}}]$ respectively. The proposed CF-Masa and Re-CA can be directly applied in the forward process of $\varepsilon_\theta([\mathcal{Z}_t^{\text{src}}, \mathcal{Z}_t^{\text{edit}}], t, [\mathcal{P}_{\text{src}}, \mathcal{P}_{\text{tgt}}])$.

\subsubsection{Cross-Frame Mutual Self-Attention}

The denoising UNet consists of different size downsample/upsample blocks and a middle block, which have four resolution levels in the latent space. Each resolution level incorporates a 2D convolution layer followed by self-attention and cross-attention layers. The attention mechanism can be formulated as:

\vspace{-0.3cm}
\begin{equation}
\label{eq:attention}
\begin{aligned}
    \text{attn}(Q, K, V) = \text{softmax}(\frac{QK^T}{\sqrt{d}})V.
\end{aligned}
\end{equation}

In self-attention layers, $Q, K, V$ are the query, key, and value features obtained by projecting the same spatial features. Without attention control, the self-attention output of source branch $\text{attn}(Q^{\text{src}}, K^{\text{src}}, V^{\text{src}})$ and editing branch $\text{attn}(Q^{\text{edit}}, K^{\text{edit}}, V^{\text{edit}})$ are computed concurrently and independently of each other. We make two changes on self-attention layers to preserve content consistency as well as temporal consistency between and within editing latent and source latent. In contrast to MasaCtrl~\cite{cao2023masactrl}, the preservation of content consistency in \OurMethod~is achieved by replacing $Q^{\text{edit}}$ and $K^{\text{edit}}$ with $Q^{\text{src}}$ and $K^{\text{src}}$ after a fixed step $t_s$ and the editing branch remains unchanged before $t_s$. To further maintain temporal consistency across batched latents within a branch, we concatenate the key features $[K_1, K_2, ..., K_m]$ and value features $[V_1, V_2, ..., V_m]$ along their sequence length dimension resulting in the final format becomes:

\vspace{-0.3cm}
\begin{equation}
\label{eq:CF-Masa}
\begin{aligned}
    &\text{CF-Masa}(\{Q_i^{\text{edit}},K_i^{\text{edit}},V_i^{\text{edit}}\}, t)  \\
    &:=\begin{cases}\{Q_i^{\text{src}},\text{concat}\{K^{\text{src}}\},\text{concat}\{V^{\text{edit}}\}\}&t \geq t_s \\
    \{Q_i^{\text{edit}},\text{concat}\{K^{\text{edit}}\},\text{concat}\{V^{\text{edit}}\}\} & t < t_s\end{cases}.
\end{aligned}
\end{equation}

\subsubsection{Re-weighted Cross Attention}

The forward process of cross-attention can be edited in a similar way to self-attention. In cross-attention layers, $Q$ is the set of query features obtained obtaining by projecting spatial features coming from self-attention layer, $K,V$ are obtained from the prompt embeddings. By replacing the cross-attention map of the editing branch with that of the source branch~\cite{hertz2022prompt}, the scattering from source prompt mutual content to the source spatial features can be maintained on editing spatial features. To further enhance the effect of the editing token, the corresponding attention map of the editing token can be multiplied by a replace scale $r \geq 1$. The resulting formulation of the Re-weighted Cross Attention is given by:

\vspace{-0.3cm}
\begin{equation}
\label{eq:Re-CA}
\begin{aligned}
    \text{Refine}(A^{\text{src}},A^{\text{edit}})_{i,j}
    = \begin{cases}
        \left(A^{\text{edit}}\right)_{i,j} 
        & \text{if} \ f_{\mathcal{P}}(j)=\text{None} \\ 
        \left(A^{\text{edit}}\right)_{i,f_{\mathcal{P}}(j)} 
        & \text{otherwise}
    \end{cases} \\
    \text{Re-CA}(A^{\text{src}},A^{\text{edit}},t)
    :=\begin{cases}
        r \cdot \text{Refine}(A^{\text{src}},A^{\text{edit}}) &t \geq t_c \\
        A^{\text{edit}} & t < t_c
    \end{cases}
\end{aligned}
\end{equation}

\noindent
where $f_{\mathcal{P}}(\cdot)$ is the alignment function indicating the source prompt token index of the $j^{th}$ token in the target prompt and \textit{None} if missing.

\subsection{Background Preservation via Latent Blending}

There is a trade-off in existing video editing methods between the editing effect of foreground objects and content preservation of background. Changing the attributes of an object in the foreground usually makes the background more consistent with the change. This is because the control methods that are applied to the forward process are not strict control over the latent space. Therefore the change of tokens in the target prompt also influences irrelevant regions of editing latent through attention mechanisms. Compared with state-of-the-art video editing methods, a significant advantage of \OurMethod~is the accuracy of foreground editing. This is shown in both quantative and qualitative results in Sec.~\ref{sec:experiments}. We achieve this by multiple designs of \OurMethod. Consistent initial latents and noise in Batch Consistency Sampling algorithm and attention control both provide faithful editing concerning the source video. In addition to this, we propose further background preservation strategies to enhance the faithfulness of the edited content to the source content. Specifically, we propose to simultaneously denoise a background branch that maintains the structure information of the editing branch while aligning content with the source branch. Based on the background branch, we additionally propose a latent blending algorithm that replaces the background part in the editing latent with the corresponding part in the background latent.

\vspace{-0.45cm}
\subsubsection{Background Branch}

By simultaneously denoising a background branch conditioned on $\mathcal{P}_{src}$ and imposing self-attention control from the source branch and editing branch, we expect the background branch to maintain the structure of the editing branch and the content of the source branch. We modify the self-attention process of the background branch as follows:

\vspace{-0.3cm}
\begin{equation}
\label{eq:bg}
\begin{aligned}
    &\text{Bg-Masa}(\{Q_i^{\text{bg}},K_i^{\text{bg}},V_i^{\text{bg}}\}, t) \\
    &:= \begin{cases}\{Q_i^{\text{src}},\text{concat}\{K^{\text{src}}\},\text{concat}\{V^{\text{src}}\}\}&t \geq t_{bg} \\
    \{Q_i^{\text{edit}},\text{concat}\{K^{\text{src}}\},\text{concat}\{V^{\text{src}}\}\} & t < t_{bg}.\end{cases}
\end{aligned}
\end{equation}

To maintain the editing structure and source content, we employ a similar editing approach to MasaCtrl~\cite{cao2023masactrl} since query features from the edit branch are used to maintain structure information. Meanwhile, the key and value features are copied from the source branch to maintain consistency with the source content. Note that the joint attention is working at early timestep instead of later timesteps as described in MasaCtrl~\cite{cao2023masactrl} because our observation is that the structure is formed at early steps and content details are refined at later steps.

\subsubsection{Latent Blending}

At the end of each denoising step, we employ the latent blending operation to replace the background region of the editing latent with the corresponding region of the source latent. The region is determined by computing the relative region from a cross-attention map. Specifically, given a cross-attention map $(A^{\text{edit}})_{m \times n}$, we obtain a blending map $(M^{\text{edit}})_m$ where $m$ is the sequence length of the attention map or the size of the feature map, and $n$ is the number of tokens in $\mathcal{P}_{tgt}$. The blending map is computed as follows:

\vspace{-0.3cm}
\begin{equation}
\label{eq:repmap}
\begin{aligned}
    (\hat{A}^{\text{edit}})_i &=
    \frac{\Sigma_j (A^{\text{edit}})_{i, j} \cdot \mathbf{I}_{f_{\mathcal{P}}(j) \neq \text{None}}}
    {\Sigma_j (A^{\text{edit}})_{i, j}} \\
    (M^{\text{edit}})_i &=
    \mathbf{I}_{(\hat{A}^{\text{edit}})_i} \geq \text{thresh}_{\text{edit}}.
\end{aligned}
\end{equation}

Intuitively, the blending map has $1$ at positions where the edited tokens receive high attention scores among all the tokens, and $0$ anywhere else. In practice, $A^{\text{edit}}$ is obtained by averaging among all the cross-attention maps of the same size in a fixed resolution level. The blended edited latent at the end of denoising step $t$ is:

\vspace{-0.3cm}
\begin{equation}
\label{eq:repedit}
\begin{aligned}
    (z_t^{\text{edit}}) = M_t^{\text{edit}} \odot (z_t^{\text{edit}})
    + (1 - M_t^{\text{edit}}) \odot (z_t^{\text{bg}}).
\end{aligned}
\end{equation}

\subsection{Frame Consistency with Tokenflow}

Following ~\cite{geyer2023tokenflow}, we apply tokenflow to improve temporal consistency across frames. Tokenflow is a plug-and-play module that can be applied at each layer of the denoising network. The idea of Tokenflow is to first select and denoise a group of keyframes, and then replace the original spatial features with the weighted sum of the two most similar spatial features from two adjacent keyframes when denoising each frame latent. In the first stage, Tokenflow selects a group of keyframes of indices $\kappa$ and in each layer at each step and store $\mathbf{T}_{base}=\{\phi(z^i)\}_{i \in \kappa}$, where $\phi(\cdot)$ maps the latent to its spatial features $(z^i)$. When computing the features of an arbitrary frame latent $z^i$, the method queries its two adjacent frames latent of indices $i-$ and $i+$, and gets the closest feature index $ \gamma^{i\pm}[p]$ for each of its feature indexed $p$ as follows:

\vspace{-0.2cm}
\begin{equation}
    \gamma^{i\pm}[p] = \argmin_{q}{\mathcal{D}\left({\phi({z}^i)[p]}, {\phi({z}^{i\pm})[q]}\right)}
\end{equation}

\noindent
where $\mathcal{D}$ represents cosine distance of two features. The output weighted spatial features of frame latent $z_i$ therefore become:

\vspace{-0.3cm}
\begin{equation}
\label{eq:tokenFlow}
\begin{aligned}
    \mathcal{F}_{\gamma}(\mathbf{T}_{base},i, p) &= w_i \cdot \phi(z^{i+}) [\gamma^{i+}[p]] \\
    &+ (1-w_i) \cdot \phi(z^{i-}) [\gamma^{i-}[p]].
\end{aligned}
\end{equation}

In general, Tokenflow is a plug-and-play operation that can be applied after the self-attention layer. It replaces the original output of spatial features $\phi(z^i)$ of the original frame latent with the features of weighted sum of two adjacent key frames $\{\mathcal{F}_{\gamma}(\mathbf{T}_{base},i, p)\}_p$.

The overall \OurMethod~ algorithm is shown in Algorithm~\ref{alg:1} and Figure~\ref{fig:overview}.

\begin{algorithm}[H]
    \begin{minipage}{\linewidth}
        \caption{\OurMethod~editing}
        \label{alg:1}
        \begin{algorithmic}[1]
            \Statex For abbreviation, we denote $\mathcal{A} \sim \mathcal{P}$ as every element in the $\mathcal{A}$ has the same value sampled from distribution $\mathcal{P}$.
            \Statex \textbf{Input:} 
            \Statex \hskip\algorithmicindent Latent Consistency Model $\varepsilon_\theta(\cdot, \cdot, \cdot)$
            \Statex \hskip\algorithmicindent Sequence of timesteps $\tau_{1}>\tau_2>\cdots >\tau_{N-1}$
            \Statex \hskip\algorithmicindent Source latents $\mathcal{Z}_0^{\text{src}} = \{z_0^{\text{src}, (i)}  \ | \ 1 \leq i \leq m \}$
            \Statex \hskip\algorithmicindent Source and target prompts $\mathcal{P}_{src}, \mathcal{P}_{tgt}$
            \State Set batch attention control on $\varepsilon_\theta(\cdot, \cdot, \cdot)$
            \State Set Tokenflow propagation on $\varepsilon_\theta(\cdot, \cdot, \cdot)$
            \State Initial batched latents $\mathcal{Z}_{\tau_1}^{\text{src}} = \mathcal{Z}_{\tau_1}^{\text{edit}} =
            \mathcal{Z}_{\tau_1}^{\text{bg}} \sim \mathcal{N}(\bm{0},\bm{I})$
            \State Compute $\{\varepsilon^{\text{cons}}_{\tau_1}\}$ using Eq \ref{eq:src}
            \For{$n=1$ to $N-1$}
                \State Compute $\mathbf{T}_{\text{base}}^{\text{edit}}$ and $\mathbf{T}_{\text{base}}^{\text{edit}}$
                \State Denoise three branches:
                \Statex \hskip\algorithmicindent $\{\varepsilon_\theta(\{z^{\text{src}}_{\tau_n}, z^{\text{edit}}_{\tau_n}, z^{\text{bg}}_{\tau_n}\}, {\tau_n}, \{\mathcal{P}_{src}, \mathcal{P}_{tgt}, \mathcal{P}_{src})\}; \mathbf{T}_{\text{base}}\}$
                \State Update $\mathcal{Z}^{\text{src}}_{\tau_{n+1}}$ using Eq \ref{eq:src}
                \State Update $\mathcal{Z}^{\text{edit}}_0$ and $\mathcal{Z}^{\text{bg}}_0$ using Eq \ref{eq:edit}
                \State Sample reconstruction noise {$\{\varepsilon_{\tau_{n+1}}^{\text{cons}}\} \sim \mathcal{N}(\bm{0},\bm{I})$}
                \State Update $\mathcal{Z}^{\text{edit}}_{\tau_{n+1}}$ and $\mathcal{Z}^{\text{bg}}_{\tau_{n+1}}$ using Eq \ref{eq:src}
                \State Replace latents $\mathcal{Z}^{\text{edit}}_{\tau_{n+1}}$ using Eq \ref{eq:repmap} and \ref{eq:repedit}
            \EndFor
            \State \textbf{Output:} $\mathcal{Z}_0^{\textrm{edit}}$
        \end{algorithmic} 
    \end{minipage}
\end{algorithm}

\section{Experiments}
\label{sec:experiments}

In this section, we first introduce the evaluation benchmark and evaluation metrics used in our experiment in Sec.~\ref{sec:evaluation_benchmark_and_metrics}. 
Following that, we present a quantitative comparison of our methods in Sec.~\ref{sec:quantitative_comparison} and a qualitative comparison in Sec.~\ref{sec:qualitative_comparison}.

\subsection{Evaluation Benchmark and Metrics}
\label{sec:evaluation_benchmark_and_metrics}

\paragraph{\textbf{Evaluation Dataset.}}
For the evaluation of video editing, we utilize the TGVE 2023 open-source dataset~\cite{wu2023cvpr} as our benchmark. 
This dataset consists of 76 videos, each containing 32 frames with a resolution of 480x480 pixels. 

\vspace{-0.3cm}
\paragraph{\textbf{Evaluation Metrics.}}

Following previous work~\cite{qi2023fatezero,geyer2023tokenflow}, we evaluate the temporal consistency of our approach by utilizing clip similarity~\cite{clip} among frames (`Tem-Con'). Additionally, we measure the frame-wise editing accuracy through two metrics. `Txt-Sim' for clip similarity between the embeddings of text and image and `Clip-Acc’ for the percentage of frames where the edited image has a higher CLIP similarity to the target prompt compared to the source prompt.
Consistent with previous research~\cite{wu2023cvpr}, we acknowledge that automated metrics can be noisy and that human evaluation is more reliable. We recognize that automatic metrics may even exhibit a lack of correlation or potentially an inverse correlation with human evaluation results. Therefore, similar to the questions used in Text Guided Video Editing Competition~\cite{wu2023cvpr}, we conduct a user study involving 20 human annotators to evaluate the 76 videos. The annotators rank each group of videos from best to worst according to four aspects: (1) preservation of essential content in the source video (`P'), (2) video generation quality (`Q'), (3) temporal consistency among frames (`C'), and (4) alignment between the text and video (`T'). 
Furthermore, as an additional evaluation metric, we measure the time consumption of editing $32$ frames' video using \OurMethod~and previous methods in both the inversion and forward processes to evaluate the speed.

\subsection{Quantitative Comparison}
\label{sec:quantitative_comparison}

In Tab.~\ref{tab:compare_TGVE} we compare \OurMethod~with two \textit{additional conditional constraints incorporating methods} Rerender~\cite{yang2023rerender} and Text2Video-Zero~\cite{khachatryan2023text2video} as well as three \textit{dual-branch methods} FateZero~\cite{qi2023fatezero}, Pix2Video~\cite{ceylan2023pix2video}, TokenFlow~\cite{geyer2023tokenflow}, RAVE~\cite{kara2024rave} and DMT~\cite{yatim2024space}.

The results demonstrate that \OurMethod~achieves state-of-the-art performance in terms of temporal consistency and per-frame editing accuracy, while significantly reducing the time required for the editing process. 
Comparatively, our method outperforms previous \textit{additional conditional constraints incorporating methods} and \textit{dual-branch methods} in terms of efficiency, delivering high-quality results in less time. 
The reduction in runtime originates from two aspects: the elimination of inversion and additional condition feature extraction, and the use of fewer sampling steps.
This highlights the effectiveness and efficiency of \OurMethod~in video editing tasks.

\begin{table*}[htbp]
\centering
\renewcommand\arraystretch{1.15}
\scalebox{0.91}{
\setlength{\tabcolsep}{0.8mm}{
\begin{tabular}{c|ccc|cccc|ccc}
\toprule
\multirow{2}{*}{Model} & \multicolumn{3}{c|}{CLIP Metrics$\uparrow$} &\multicolumn{4}{c|}{User Study$\downarrow$} &  \multicolumn{3}{c}{Time$\downarrow$} \\ 
\cline{2-11}  
 & Tem-Con & Txt-Sim & Clip-Acc & P & Q & C & T & Inversion  & Forward  & Sum \\ \midrule
Rerender~\cite{yang2023rerender} & 95.7 & 25.0 & 48.5 & 5.2 & 5.8 & 5.2 & 4.6 & - & 174.3 & 174.3 \\
Text2Video-Zero~\cite{khachatryan2023text2video} & \textbf{96.9} & 27.1 & 70.7 & 7.1 & 6.7 & 4.9 & 4.9 & - & 131.0 & \textbf{131.0} \\ \midrule
FateZero~\cite{qi2023fatezero} & 95.7 & 24.9 & 35.8 & 4.5 & 4.6 & \textbf{2.9} & 4.8 & 233.7 & 347.0 & 581.7 \\
Pix2Video~\cite{ceylan2023pix2video} & 96.0 & \underline{27.5} & 68.5 & 4.6 & 4.8 & 4.8 & \underline{4.1} &  185.3 & 213.0 & 399.3 \\
TokenFlow~\cite{geyer2023tokenflow} & \underline{96.5} & 25.5 & 54.7 & 4.1 & \underline{3.2} & 4.3 & 4.8 & 176.5 & \textbf{115.9} &  292.4 \\
RAVE~\cite{kara2024rave} & 95.5 & 26.2 & 56.8 & 5.2 & 4.2 & 6.5 & 4.7 & \textbf{69.8} & 126.4 & 196.2 \\
DMT~\cite{yatim2024space} & 96.2 & 26.9 & \underline{70.9} & \textbf{2.4} & \textbf{2.9} & 3.8 & 4.2 & \textbf{44.2} & 363.5 & 407.5 \\
Ours & \underline{96.5} & \textbf{27.7} & \textbf{71.1} & \underline{2.9} & 3.8 & \underline{3.6} & \textbf{3.9} & - & \textbf{61.7} & \textbf{61.7} \\ \bottomrule
\end{tabular}}}
\small
\caption{\textbf{Comparison of \OurMethod~with previous video editing methods.} \textbf{Bold} indicates best. \underline{Underline} indicates second best.}
\label{tab:compare_TGVE}
\end{table*}

\subsection{Qualitative Comparison}
\label{sec:qualitative_comparison}

\begin{figure}[htbp]
    \centering
    \includegraphics[width=0.99\linewidth]{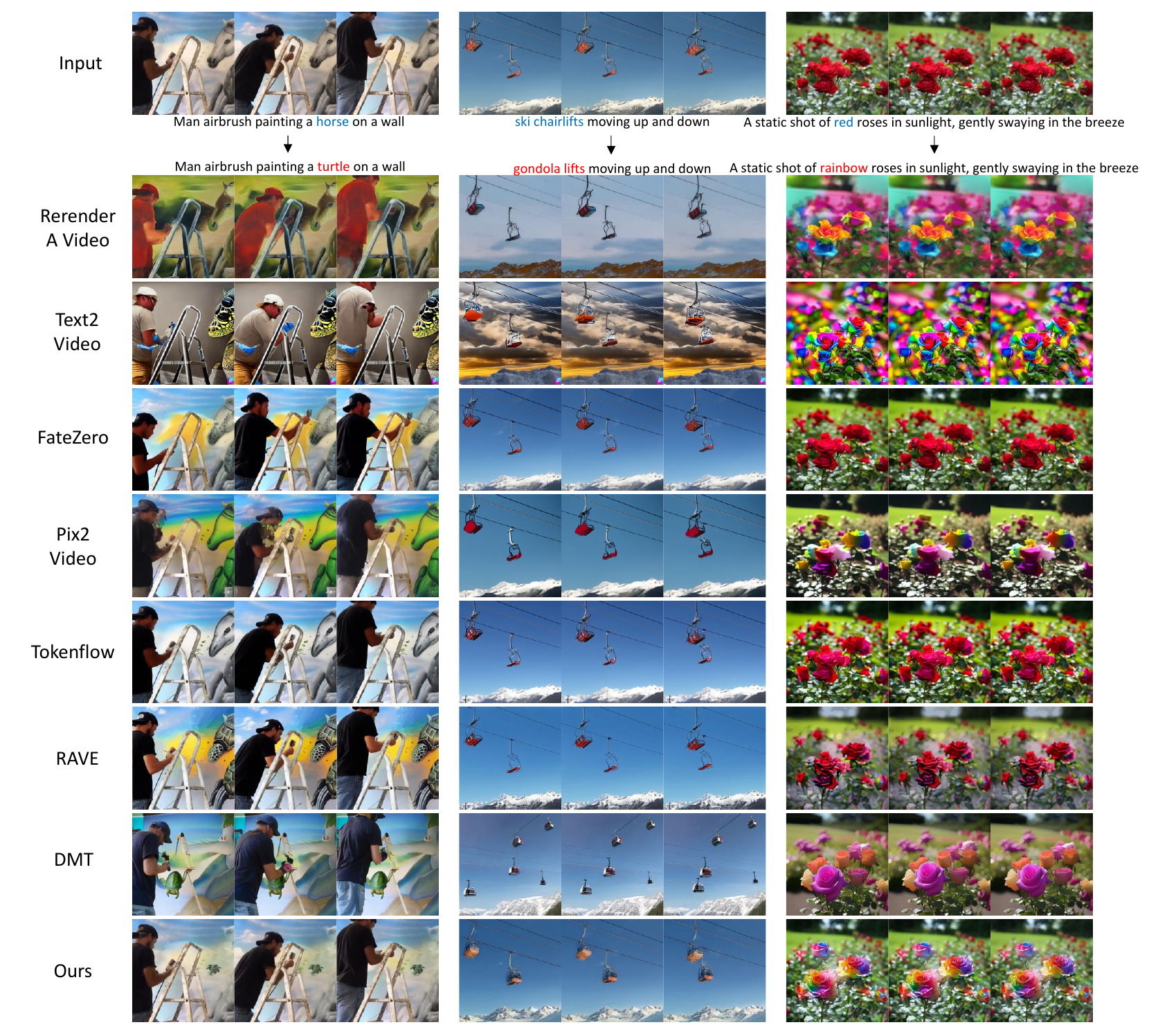}
    \caption{\textbf{Qualitative comparison of \OurMethod~with previous video editing methods.} The top row displays the source video, while the following rows showcase edited videos by previous editing methods and \OurMethod. Source and target text prompt at shown the top, with the edited words highlighted in red.
    }
    \label{fig:qualitative_results}
\end{figure}

Qualitative comparison of \OurMethod~and previous video editing methods is shown in Fig.~\ref{fig:qualitative_results}. We compare  \textit{additional conditional constraints incorporating methods} Rerender~\cite{yang2023rerender} and Text2Video-Zero~\cite{khachatryan2023text2video} as well as three \textit{dual-branch methods} FateZero~\cite{qi2023fatezero}, Pix2Video~\cite{ceylan2023pix2video}, and TokenFlow~\cite{geyer2023tokenflow}.

The results show that \OurMethod~effectively performs video editing aligned with the text prompt while preserving the essential content of the source video. Through attention control, latent blending, and leveraging the preservation ability of the consistency model, \OurMethod~successfully performs video foreground editing while preserving the background. This approach enables targeted editing of the foreground elements in the video while ensuring that the background remains intact. By selectively focusing on specific regions of interest and employing latent blending techniques, \OurMethod~achieves accurate and consistent editing results, maintaining the integrity of the background content.
It is worth noting that \OurMethod~achieves superior performance compared to other methods while requiring significantly less time. This highlights the efficiency and effectiveness of our approach in delivering high-quality results in a more time-efficient manner.

\subsection{Ablation Study}

\begin{figure}[htbp]
    \centering
    \includegraphics[width=0.8\linewidth]{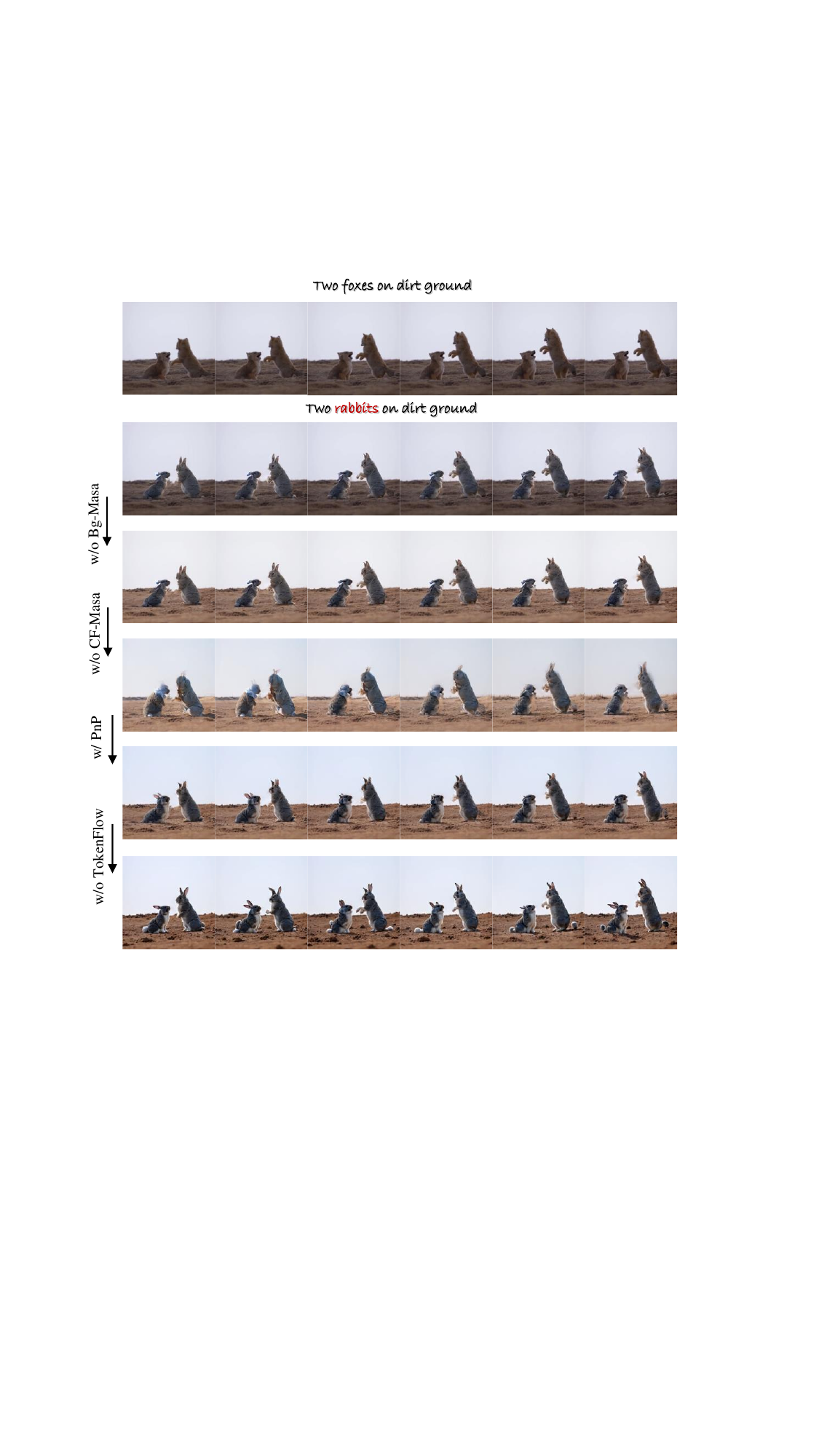}
    \caption{\textbf{Illustration of ablation on model architecture.}}
    \label{fig:ablatioin}
\end{figure}

We ablate the use of $\text{Bg-Masa}$, $\text{CF-Masa}$, $\text{Re-CA}$ and TokenFlow propagation. Quantitative results and qualitative results are shown in Tab.~\ref{tab:ablation} and Fig.~\ref{fig:ablatioin}. Without background preservation, the background dirt is changed. Results show that removing $\text{CF-Masa}$ and TokenFlow results in a worse temporal consistency. Moreover, replacing our attention control with $\text{PnP}$ results in a worse editing effect (See left rabbit's ears and right rabbit's tail).

\begin{table}[htbp]
\centering
     \renewcommand\arraystretch{1.15}
     \scalebox{0.91}{
\setlength{\tabcolsep}{0.8mm}{
\begin{tabular}{c|ccc}
\toprule
\multirow{2}{*}{Model} & \multicolumn{3}{c}{CLIP Metrics$\uparrow$}  \\ 
\cline{2-4}  
 & Tem-Con & Txt-Sim & Clip-Acc  \\ \midrule
Ours & \underline{96.5} & \textbf{27.7} & \underline{71.1} \\
w/o Bg-Masa & \textbf{96.7} & \underline{27.5} & \textbf{72.3} \\
w/o CF-Masa & 96.3 & 26.7 & 69.3  \\ 
w/ PnP & \underline{96.5} & 25.8 & 60.0  \\ \bottomrule
\end{tabular}}}
\small
\caption{\textbf{Ablation study for architecture design of \OurMethod.} \textbf{Bold} indicates the best. \underline{Underline} indicates the second best.}
\label{tab:ablation}

\end{table}

Tab.~\ref{tab:ablation} shows that without latent blending the temporal consistency and CLIP accuracy metrics rise, which illustrates that latent blending protects background but does not help with either temporal consistency or CLIP accuracy. The improvement in background preservation is observed evidently in qualitative results which is not reflected on CLIP metrics. Imposing background preservation prevents the adaption of background to the editing prompt which is negatively reflected on CLIP based similarity evaluation. However, visual observation by eyes can hardly capture the negative impact it causes in terms of content editing. Apart from background preservation designs, the rest of our proposed attention controls achieve better performance in all the three metrics, which shows the effectiveness of our proposed methods.

\section{Conclusion}
\label{sec:conclusion}

\paragraph{\textbf{Conclusion.}} In this work, we introduce \OurMethod, a zero-shot video editing approach that addresses the computational challenges faced by previous methods. By leveraging the self-consistency property of Consistency Models, our method eliminates the need for time-consuming inversion or additional condition extraction steps. We have also introduced a novel approach for maintaining background preservation via latent blending, which simultaneously denoises a background branch while imposing self-attention control from the source and editing branches.
Experimental results demonstrate the superior performance of \OurMethod~in terms of editing quality while requiring significantly less time for video editing tasks. 

\vspace{-0.3cm}
\paragraph{\textbf{Limitations and future work.}} \OurMethod~still has some limitations: (1) \OurMethod~may require tuning its hyperparameters to achieve optimal performance on each video. This dependency on hyperparameter adjustment adds complexity to the editing process and may require expertise or extensive experimentation to achieve satisfactory results. (2) While \OurMethod~demonstrates state-of-the-art performance in video editing, there is no guarantee of success for every editing case. The effectiveness of the approach may vary depending on factors such as input data quality and the complexity of the editing task.

\newpage
{\small
\bibliographystyle{ieee_fullname}
\bibliography{egbib}
}

\end{document}